\documentclass[10pt,twocolumn,letterpaper]{article}

\usepackage[pagenumbers]{cvpr} 

\usepackage{graphicx}
\usepackage{amsmath}
\usepackage{amssymb}
\usepackage{booktabs}

\usepackage{enumitem}
\setlist{leftmargin=3mm}
\usepackage{kotex}
\usepackage{caption}
\usepackage{subcaption}
\usepackage{color, colortbl}
\usepackage[dvipsnames]{xcolor}
\usepackage{algorithm}
\usepackage{algpseudocode} 
\usepackage{wrapfig}
\usepackage[accsupp]{axessibility}
\usepackage{hyperref}
\hypersetup{colorlinks,pagebackref,breaklinks}

\usepackage[capitalize]{cleveref}
\crefname{section}{Sec.}{Secs.}
\Crefname{section}{Section}{Sections}
\Crefname{table}{Table}{Tables}
\crefname{table}{Tab.}{Tabs.}

\newcommand{\our}{CMO\xspace}  
\newcommand{\ourfull}{Context-rich Minority Oversampling\xspace} 

\definecolor{sdcolor}{rgb}{0.12, 0.39, 0.60}
\definecolor{newgreen}{rgb}{0, 0.65, 0.3}

\definecolor{lightcolor}{HTML}{ededed}
\definecolor{commentcolor}{RGB}{75,125,113}   
\newcommand{\PyComment}[1]{\ttfamily\textcolor{commentcolor}{\# #1}}  
\newcommand{\PyCode}[1]{\ttfamily\textcolor{black}{#1}} 

\def\cvprPaperID{9746} 
\def\confName{CVPR}
\def\confYear{2022}

\begin{document}

\title{The Majority Can Help the Minority: \\Context-rich Minority Oversampling for Long-tailed Classification}

\author{Seulki Park$^1$\thanks{The work for this paper was performed as part of an internship at NAVER AI Lab.}\quad Youngkyu Hong$^{2}$ \quad Byeongho Heo$^2$ \quad Sangdoo Yun$^{2}$ \quad Jin Young Choi$^1$\\
{\hspace{1cm}$^1$ASRI, ECE., Seoul National University}
{\hspace{1cm}$^2$NAVER AI Lab}\\
{\tt\footnotesize {seulki.park@snu.ac.kr, yk.hong@kaist.ac.kr, \{bh.heo, sangdoo.yun\}@navercorp.com,
jychoi@snu.ac.kr}}
}

\maketitle

\begin{abstract}
The problem of class imbalanced data is that the generalization performance of the classifier deteriorates due to the lack of data from minority classes.
In this paper, we propose a novel minority over-sampling method to augment diversified minority samples by leveraging the rich context of the majority classes as background images. 
To diversify the minority samples, our key idea is to paste an image from a minority class onto rich-context images from a majority class, using them as background images. 
Our method is simple and can be easily combined with the existing long-tailed recognition methods.
We empirically prove the effectiveness of the proposed oversampling method through extensive experiments and ablation studies.
Without any architectural changes or complex algorithms, our method achieves state-of-the-art performance on various long-tailed classification benchmarks.
Our code is made available at {\url{ https://github.com/naver-ai/cmo}}.

\end{abstract}

\section{Introduction}
\label{sec:intro}

\begin{figure}[t]
\centering{%
\includegraphics[width=\columnwidth]{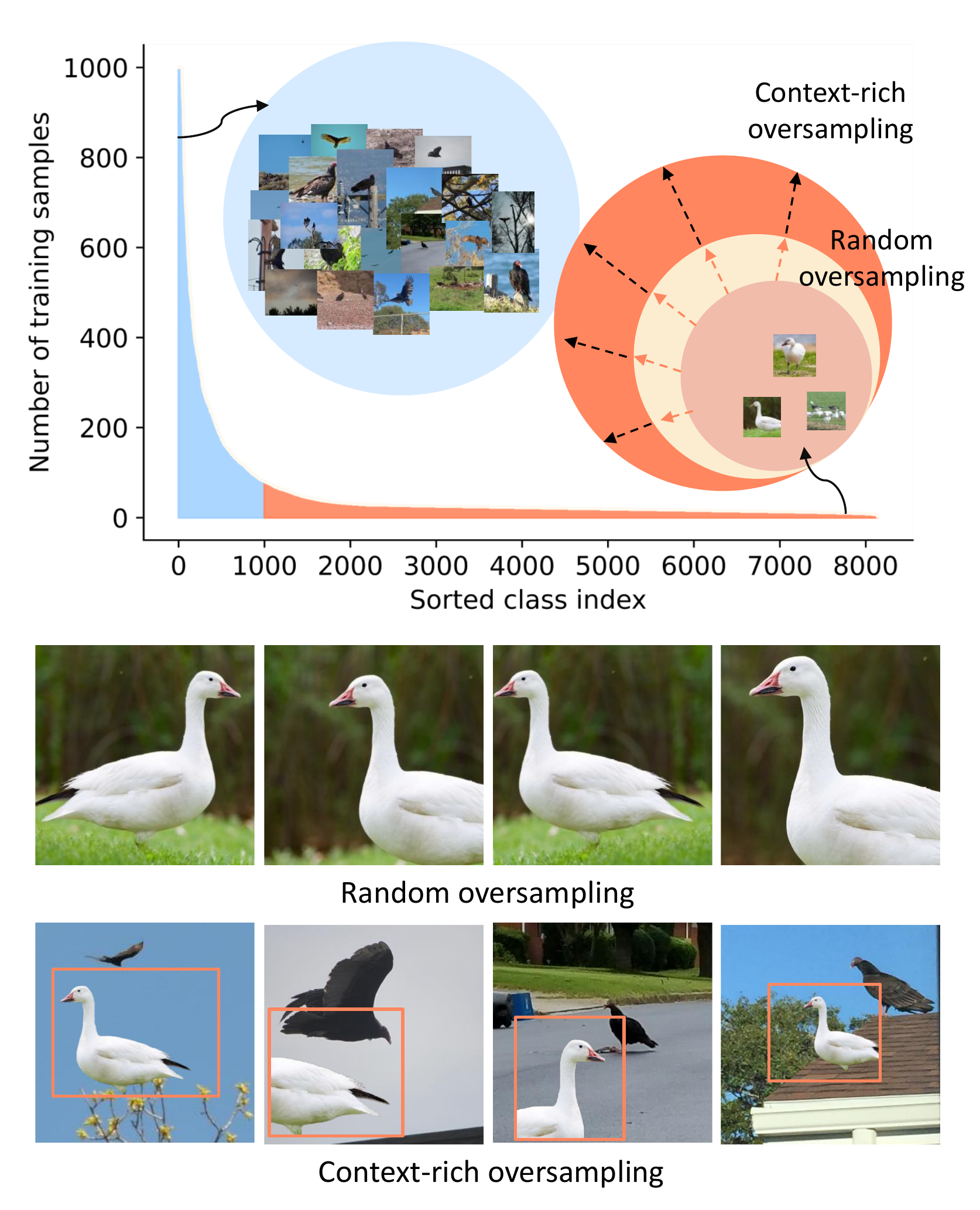}}
\caption{\textbf{Concept of context-rich minority oversampling}.
In the real-world long-tailed dataset iNaturalist 2018~\cite{ref:data_iNat}, the number of samples from the head class and the tail class is extremely different (Upper).
Simple random oversampling method repeatedly produces context-limited images from minority classes.
We propose a novel context-rich oversampling method to generate diversified minority images.
To this end, we oversample the tail-class images with various sizes. Then, these patches are pasted onto the head-class images to have various backgrounds.
Our key idea is to bring rich contexts from majority samples into minority samples.}
\label{fig:concept}
\end{figure}

Real-world data are likely to be inherently imbalanced~\cite{ref:data_celeba, ref:data_coco, ref:data_uci, ref:data_iNat}, where the number of samples per class differs greatly. 
If models are trained on an imbalanced dataset, they can be easily biased toward majority classes and tend to have a poor generalization ability on recognizing minority classes (\emph{i.e.,} overfitting).

A simple and straightforward method to overcome the class imbalance problem is to repeatedly oversample the minority classes~\cite{ref:resample_chawla_smote02,ref:resample_hulse_icml07}. 
However, these naive oversampling can intensify the overfitting problem, since the repeatedly selected samples have less diversity but almost similar image contexts~\cite{ref:sarafianos_eccv2018}. 
For example, consider a minority class of `snow goose,' in which the geese always stand upon grass in the training images.
If samples are drawn from these limited training samples~\cite{ref:resample_hulse_icml07} or even if new samples are produced by interpolating within the class~\cite{ref:resample_chawla_smote02}, only \textbf{context-limited} images will be created as in Figure~\ref{fig:concept}.
Our goal is to solve the aforementioned problem by introducing a simple \textbf{context-rich} oversampling method.

We pay attention to the characteristics of long-tailed distributions; that is, majority class samples are data-rich and information-rich.
Unlike the existing re-sampling methods that ignore (\emph{i.e.}, undersample) majority samples, our method uses the affluent information of the majority samples to generate new minority samples.
Specifically, our idea is to leverage the rich major-class images as the background for the newly created minor-class images.
Figure \ref{fig:concept} illustrates the concept of our proposed context-rich oversampling strategy.
Given an original image from a minority class, the object is cropped in various sizes and pasted onto the various images from majority classes. 
Then, we can create images with more diverse contexts (\emph{e.g.}, `snow goose' images with the sky, road, roof, crows, etc).
Since this is an interpolation of the majority and minority class samples, it generates diversified data around the decision boundary, and as a result, it improves the generalization performance for minority classes.

To this end, we adopt an image-mixing data augmentation method, CutMix~\cite{ref:cutmix_iccv2019}. 
As our key idea is to transfer rich contexts from majority to minority samples, we apply a simple and effective data sampling method to generate new minority-centric images with majority's contexts. 
However, naive use of CutMix may exacerbate the overfitting problem in favor of the majority classes because it may generate more majority-centric samples than minority samples. 
We solve this problem by sampling the background images and the foreground patches from different distributions to achieve the desired minority oversampling.

Our key contributions can be summarized as follows:
(1) We propose a novel \MakeLowercase{\ourfull} method that generates various samples by leveraging the rich context of the majority classes as background images.
(2) Our method requires little additional computational cost and can be easily integrated into many end-to-end deep learning algorithms for long-tailed recognition.
(3) We demonstrate that significant performance improvements and state-of-the-art performance can be achieved by applying the proposed oversampling to existing commonly used loss functions without any architectural changes or complex algorithms. 
(4) We empirically prove the effectiveness of the proposed oversampling method through extensive experiments and ablation studies.
We believe that our study offers a useful and universal minority oversampling method for research into long-tailed classification.

\section{Related Work}
\label{sec:relatedwork}

\subsection{Long-tailed Recognition}
\textbf{Re-weighting methods.}
Re-weighting aims to assign different weights to training samples to adjust their importance either at the class level or at the instance level.
Class-level re-weighting methods include re-weighting samples by inverse class frequency~\cite{ref:huang_tang_cvpr16, ref:wang_hebert_nips17}, Class-balanced loss~\cite{ref:cui_belongie_cvpr19}, LDAM loss~\cite{ref:cao_ldam_neurips2019}, Balanced Softmax~\cite{ref:ren_bms_neurips2020}, LADE loss~\cite{ref:lade_cvpr2021}.
Instance-level re-weighting methods include focal loss~\cite{ref:lin_focal_loss_iccv17} and influence-balanced loss~\cite{ref:ibloss_iccv2021}.

\textbf{Re-sampling methods.}
Resampling methods aim to modify the training distributions to decrease the level of imbalance~\cite{ref:johnson_imbal_survey_2019}.
Resampling methods include undersampling and oversampling.
Undersampling methods~\cite{ref:zhang_resample_icml2003, ref:resample_hulse_icml07} that discard the majority samples can lose valuable information, and undersampling is infeasible when the imbalance between classes is too high.
The simplest form of oversampling is random oversampling (ROS)~\cite{ref:resample_hulse_icml07, ref:buda_mazurowski_2018}, which oversamples all minority classes until class balance is achieved.
This method is simple and can be easily used in any algorithm, but since the same sample is repeatedly drawn, it can lead to overfitting~\cite{ref:sarafianos_eccv2018}.
As a more advanced method, the synthetic minority over-sampling technique (SMOTE)~\cite{ref:resample_chawla_smote02}, which oversamples minority samples by interpolating between existing minority samples and their nearest minority neighbors, was proposed.
Following the success of SMOTE, several variants have been developed: Borderline-SMOTE~\cite{ref:borderlinesmote_2005}, which oversamples the minority samples near class borders, and Safe-level-SMOTE~\cite{ref:safesmote_2009}, which defines safe regions not to oversample samples from different classes.
These methods have been widely used by classical machine learning algorithms, but there are difficulties in using them for large-scale image datasets due to the high computational complexity of calculating the K-Nearest Neighbor for every sample.
Generative adversarial minority oversampling (GAMO)~\cite{ref:sankha_gamo_iccv2019} solves this issue by producing new minority samples by training a convex generator, inspired by the success of generative adversarial networks (GANs)~\cite{ref:gan_nips14} in image generation. 
However, training the generator incurs high additional training cost; moreover, GAMO can suffer from the infamous mode collapse of GANs~\cite{ref:Bau_2019_ICCV}.
To generate diverse minority data, recent works~\cite{ref:m2m_cvpr2020, ref:blt_accv2020} have proposed adversarial augmentations by adding small noise to the input images.
To this end, Major-to-minor Translation (M2m)~\cite{ref:m2m_cvpr2020} transfers knowledge from majority classes using a pre-trained network, and Balancing Long-Tailed datasets (BLT)~\cite{ref:blt_accv2020} uses a gradient-ascent image generator based on the confusion matrix. 

Another recent line of research is oversampling in the feature space rather than in the input space: Deep Over-sampling (DOS)~\cite{ref:shin_dos_2017}, Feature-space Augmentation (FSA)\cite{ref:feature_aug_eccv2020}, and Meta Semantic Augmentation (MetaSAug)~\cite{ref:metasaug_cvpr2021}.
These methods aim to augment minority classes in the feature space by sampling from the in-class neighbors in the linear subspace ~\cite{ref:shin_dos_2017}, using learned features from pretrained networks~\cite{ref:feature_aug_eccv2020}, or using an implicit semantic data augmentation (ISDA) algorithm~\cite{ref:isda_neurips2019}.
However, DOS~\cite{ref:shin_dos_2017} requires finding the nearest neighbors in the feature space, FSA~\cite{ref:feature_aug_eccv2020} requires a pre-trained feature sub-network and a classifier for feature augmentation procedure.
Lastly, MetaSAug~\cite{ref:metasaug_cvpr2021} demands additional uniform validation samples that outnumber the number of samples in the tail classes and hundreds and thousands of iterations for training. 
Consequently, these methods are less cost-efficient and technically more difficult to perform.
On the other hand, our method oversamples diverse minority samples using a simple data augmentation technique and outperforms all previous methods while maintaining reasonable training costs.

\textbf{Other long-tailed methods.}
Recently, significant improvement has been achieved by two-stage algorithms: Deferred re-weighting (DRW)~\cite{ref:cao_ldam_neurips2019}, classifier re-training (cRT), learnable weight scaling (LWS)~\cite{ref:kang_decoupling_iclr2020}, and the Mixup shifted label-aware smoothing model (MiSLAS)~\cite{ref:mislas_cvpr2021}.
Meanwhile, a bilateral branch network (BBN)~\cite{ref:zhou_chen_bbn_cvpr2020} uses an additional network branch for re-balancing, and RIDE~\cite{ref:ride_iclr2021} uses multiple branches called experts, each of which learns to specialize in different classes.
Another line of recent research employs meta-learning methods: Meta-Weight-Net~\cite{ref:shu_metaweightnet_neurips2019} learns an explicit loss-weight function, and a meta sampler~\cite{ref:ren_bms_neurips2020} estimates the optimal class sample rate. 
PaCo~\cite{ref:paco_iccv2021} proposes supervised contrastive learning with parametric class-wise centers for long-tailed classification.
\vspace{-2mm}
\subsection{Data Augmentation and Mixup Methods}
\vspace{-1mm}
Spatial-level augmentation methods have performed satisfactorily in the computer vision fields.
Cutout~\cite{ref:cutout_2017} removes random regions whereas CutMix~\cite{ref:cutmix_iccv2019} fills the removed regions with patches from another training image.
In addition, mixup methods~\cite{ref:mixup_iclr2018, ref:manifoldmixup_2019, ref:improvedmixup_2019} linearly interpolate two images in a training dataset.
Since the data augmentation method is closely related to the oversampling methods, some recent long-tailed recognition methods have used the mixup method.
Zhou et al.~\cite{ref:zhou_chen_bbn_cvpr2020} use the mixup as a baseline method, and MiSLAS~\cite{ref:mislas_cvpr2021} uses mixup in its Stage-1 training.
However, these methods apply mixup without any adjustments, and little work has been done to explore appropriate data augmentation techniques for a long-tailed dataset.
Recently, for an imbalanced dataset, the Remix~\cite{ref:remix_eccv2020} assigned a label in favor of the minority classes when mixing two samples. 
Unlike these methods, our method samples images from different distributions, which takes into account the specificity of long-tailed data distribution.

\vspace{-1mm}
\section{\ourfull}
\label{sec:method}

\subsection{Algorithm}
\label{sub:method_algorithm}
We propose a new oversampling method called \ourfull (\our).
\our utilizes the contexts of the majority samples to diversify the limited context of the minority samples.
As shown in the Figure~\ref{fig:concept}, the background images are sampled from majority classes and combined with foreground images of minority classes.
Let $x \in \mathbb{R}^{W \times H \times C}$ and $y$ denote a training image and its label, respectively.
We aim to generate a new sample $(\Tilde{x}, \Tilde{y})$ by combining two training samples $(x^b, y^b)$ and $(x^f, y^f)$.
Here, the image $x^b$ is used as a background image, and the image $x^f$ provides the foreground patch to be pasted onto $(x^b, y^b)$.

For the image combining method, we chose CutMix~\cite{ref:cutmix_iccv2019} data augmentation due to its simplicity and effectiveness.
Following CutMix~\cite{ref:cutmix_iccv2019} settings, the image and label pairs are augmented as
\begin{align}
    \Tilde{x} &= \textbf{M} \odot x^b + (\textbf{1} - \textbf{M}) \odot x^f \nonumber \\
    \Tilde{y} &= \lambda y^b + (1 - \lambda) y^f,
    \label{eq:cutmix}
\end{align}
where $(\textbf{1} - \textbf{M})\in \{0, 1\}^{W \times H}$ denotes a binary mask indicating where to select the patch and paste it onto a background image.
$\textbf{1}$ means a binary mask filled with ones, and $\odot$ is element-wise multiplication.
The combination ratio $\lambda\in\mathbb{R}$ between two images is sampled from the beta distribution $Beta(\alpha, \alpha)$.
To sample the mask and its coordinates, we apply the original CutMix~\cite{ref:cutmix_iccv2019} setting.
An experiment on using a different $\alpha$ is included in the Supplementary Material.

Since CutMix was originally designed for data augmentation on a class-balanced dataset, Eq.~\ref{eq:cutmix} does not represent the majority or minority class of samples.
To change the method to \our, we include sampling data distributions for foreground $(x_f, y_f)$ and background samples $(x^b, y^b)$.
In our design, the background samples $(x^b, y^b)$ should be biased to the majority classes.
Therefore, we sample the background samples from the original data distribution $P$.
Meanwhile, the foreground samples $(x^f,y^f)$ are sampled from minor-class-weighted distribution $Q$ to be biased to the minority classes.
In short, \our consists of data sampling from two distributions, $(x^b, y^b) \sim P$ and $(x^f, y^f) \sim Q $, and combining the images using Eq.~\ref{eq:cutmix}.
The pseudo-code of the training procedure is presented in Algorithm~\ref{alg:full_method}.
\vspace{-3mm}




\label{app:algo}
\begin{algorithm}[h!]
    \caption{\ourfull (\our)} 
	\label{alg:full_method}
	\begin{algorithmic}[1]
	    \Require Dataset $\mathcal{D}_{i=1}^N$, model parameters  $\theta$, $P$, $Q$, any loss function $L(\cdot)$.
	    \State {Randomly initialize $\theta$.}
       \State{Sample weighted dataset $\Tilde{\mathcal{D}}_{i=1}^N \sim Q$}.
	    \For {epoch $= 1, \ldots, T$}
	       \For {batch $i= 1, \ldots, B$}
	            \State{Draw a mini-batch $(x^b_i, y^b_i)$ from $\mathcal{D}_{i=1}^N$}
	            \State{Draw a mini-batch $(x^f_i, y^f_i)$ from $\Tilde{\mathcal{D}}_{i=1}^N$} 
	           \State{$\lambda \sim Beta(\alpha, \alpha)$}
	           \State{$\Tilde{x}_i = \textbf{M} \odot x^b_i + (\textbf{1} - \textbf{M}) \odot x^f_i$}
	           \State{$\Tilde{y}_i = \lambda y^b_i + (1 - \lambda) y^f_i$}
	           \State{$\theta \leftarrow \theta -\eta \nabla L((\Tilde{x}_i, \Tilde{y}_i);\theta)$}
	       \EndFor
	    \EndFor
	\end{algorithmic} 
\end{algorithm}
\vspace*{-5mm}
\vspace{-1mm}
\subsection{Minor-class-weighted Distribution $Q$}
\label{sub:method_q}
\vspace{-1mm}
To sample the foreground image from minority classes, we design the minor-class-weighted distribution $Q$ by utilizing the re-weighting methods.
The re-weighting approach, dating back to the classical importance sampling method~\cite{ref:importance_sampling_1953}, provided a way to assign appropriate weights to samples.
Commonly used sampling strategies include ones that assign a weight inversely proportional to the class frequency~\cite{ref:huang_tang_cvpr16, ref:wang_hebert_nips17}, to the smoothed class frequency~\cite{ref:mikolov_nips2013, ref:Mahajan_Weiss_eccv2018}, or to the effective number~\cite{ref:cui_belongie_cvpr19}.

Let $n_k$ be the number of samples in the $k$-th class, then for the $C$ classes, the total number of samples is $N = \sum_{k=1}^C n_k$.
Then, the generalized sampling probability for the $k$-th class can be defined by 
\vspace{-2mm}
\begin{equation}
    q(r, k) = \frac{1 / n^r_k}{\sum_{k^{'}=1}^{C} 1 / n^r_{k^{'}}},
\end{equation}
where the $k$-th class has a sampling weight inversely proportional to $n^r_k$.
As $r$ increases, the weight of the minor class becomes increasingly larger than that of the major class.
By adjusting the value of $r$, we can examine diverse sampling strategies.
Setting $r=1$ uses the inverse class frequency~\cite{ref:huang_tang_cvpr16, ref:wang_hebert_nips17} while setting $r=1/2$ uses the smoothed inverse class frequency, as in~\cite{ref:mikolov_nips2013, ref:Mahajan_Weiss_eccv2018}.
We can also use the effective number~\cite{ref:cui_belongie_cvpr19} instead of $n^r_k$, which is defined as
\begin{equation}
    E(k) = \frac{(1 - \beta^{n_k})}{(1-\beta)}, 
\end{equation}
where $\beta = (N-1)/N$.
Since \our is a new approach for long-tailed classification, it is hard to predict the performance of each sampling strategy for \our. Therefore, we evaluate the different sampling strategies on the long-tailed CIFAR-100~\cite{ref:data_cifar} and select the best strategy $q(1,k)$ for the minor-class-weighted distribution $Q$. The experimental results are displayed in Table~\ref{table:res-dist-cifar100} of the experimental section.

\subsection{Regularization Effect of \our}
\label{sub:method_reg}
A recent study~\cite{ref:mislas_cvpr2021} has reported that models trained on long-tailed datasets are more over-confident than the models trained on balanced data.
In addition, the study reveals that the long-tailed classification accuracy can be improved by solving the over-confidence issue.
Moreover, \our can be interpreted as a way to mitigate over-confidence in long-tailed classification.
Inherited from CutMix, \our uses a soft-target label $\Tilde{y}$, as in Eq.~\ref{eq:cutmix}.
The soft-target label penalizes over-confident outputs, similarly to the label smoothing regularization~\cite{ref:labelsmoothing_cvpr2016}.
Therefore, we argue that \our contributes not only to minority sample generation but also to mitigating the over-confidence, which both enable an impressive performance improvement in diverse long-tail settings.
We will demonstrate the effectiveness of \our through various experiments in the experimental section.
\section{Experiments}
\label{sec:experiment}

We present experiments on and analyses of \our in this section. 
We first describe our experimental settings and implementation details in Section~\ref{sub:exp_setting}.
Next, we present the effectiveness of \our using three long-tailed classification benchmarks: CIFAR-100-LT, ImageNet-LT, and iNaturalist. \our consistently boosts the performance of these baselines with state-of-the-art accuracy (Section~\ref{sub:exp:main_exp}). 
In Section~\ref{sub:exp_analysis} we present in-depth analyses of \our to study its inherent characteristics.


\subsection{Experimental Settings} \label{sub:exp_setting}
\noindent\textbf{Datasets.} 
We validate \our on the most commonly used long-tailed recognition benchmark datasets: CIFAR-100-LT\cite{ref:cao_ldam_neurips2019}, ImageNet-LT~\cite{ref:liu_imagenetlt_cvpr2019}, and iNaturalist 2018~\cite{ref:data_iNat} (see Table~\ref{table:dataset}).
CIFAR-100-LT and ImageNet-LT are artificially made imbalanced from their balanced versions (CIFAR-100~\cite{ref:data_cifar} and ImageNet-2012~\cite{ref:data_imagenet}).
The iNaturalist 2018 dataset is a large-scale real-world dataset that exhibits long-tailed imbalance.
We used the official training and test splits in our experiments.
\begin{table}[h]
\vspace{0mm}
\caption{
\textbf{Summary of datasets}. The imbalance ratio $\rho$ is defined by $\rho = \max_k \{n_k\}/{\min_k \{n_k\}}$, where $n_k$ is the number of samples in the $k$-th class.
}
\label{table:dataset}
\centering
\tabcolsep=0.1cm
\small
{
\begin{tabular}{lccc}
\toprule
Dataset          & \# of classes & \# of training & Imbalance ratio \\ \midrule
CIFAR-100-LT     & 100           & 50K            &\{10, 50, 100\}     \\
ImageNet-LT      & 1,000         & 115.8K         & 256             \\
iNaturalist 2018 & 8,142         & 437.5K         & 500             \\ \bottomrule
\end{tabular}
}
\end{table}
\vspace{0mm}

\noindent\textbf{Evaluation Metrics.} 
Performances is mainly reported as the overall top-1 accuracy. 
Following~\cite{ref:liu_imagenetlt_cvpr2019}, we also report the accuracy of three disjoint subsets: Many-shot classes (classes that contain more than 100 training samples), medium-shot classes (classes that contain 20 to 100 samples), and few-shot classes (classes that contain under 20 samples).

\noindent\textbf{Comparison methods.}
We compare \our with the minority oversampling methods, the state-of-the-art long-tail recognition methods, and their combinations. 
\begin{itemize}
\item \textbf{Minority oversampling.} 
(1) No oversampling (vanilla); 
(2) Random oversampling (ROS)~\cite{ref:resample_hulse_icml07}, that oversamples minority samples to balance the classes in the training data;
(3) Remix~\cite{ref:remix_eccv2020}, which oversamples minority classes by assigning higher weights to the minority labels when using Mixup~\cite{ref:mixup_iclr2018};
(4) Feature space augmentation (FSA)~\cite{ref:feature_aug_eccv2020}.
\vspace{-1mm}
\item \textbf{Re-weighting.}
(5) label-distribution-aware margin (LDAM) loss~\cite{ref:cao_ldam_neurips2019}, which regularizes the minority classes to increase margins to the decision boundary;
(6) influence-balanced (IB) loss~\cite{ref:ibloss_iccv2021}, which re-weights samples by their influences;
(7) Balanced Softmax~\cite{ref:ren_bms_neurips2020}, an unbiased extension of Softmax;
(8) LADE~\cite{ref:lade_cvpr2021}, which disentangles the source label distribution from the model prediction. 
\vspace{-1mm}
\item \textbf{Other state-of-the-art methods.}
(9) Deferred re-weighting (DRW)~\cite{ref:cao_ldam_neurips2019} and (10) Decouple~\cite{ref:kang_decoupling_iclr2020} are two-stage algorithms that re-balance the classifiers during fine-tuning;
(11) BBN~\cite{ref:zhou_chen_bbn_cvpr2020} and (12) RIDE~\cite{ref:ride_iclr2021} use additional network branches to handle class imbalance;
(13) Causal Norm~\cite{ref:causalnorm_neurips2020}, which disentangles causal effects and adjusts the effects in training;
(14) MiSLAS~\cite{ref:mislas_cvpr2021}, a two-stage algorithm, enhances classifier learning and calibration with label-aware smoothing (LAS) in stage-2.
\end{itemize}

\noindent\textbf{Implementation.} \label{sub:implementation}
We use PyTorch \cite{ref:Pytorch} for all experiments.
For the CIFAR datasets, we use ResNet-32~\cite{ref:resnet_cvpr2016}.
The networks are trained for 200 epochs following the training strategy in \cite{ref:cao_ldam_neurips2019}.
For ImageNet-LT, we use ResNet-50 as the backbone network.
The network is trained for 100 epochs using an initial learning rate of 0.1.
The learning rate is decayed at the 60th and 80th epochs by 0.1.
For iNaturalist 2018, we use ResNet-\{50, 101, 152\} and Wide ResNet-50~\cite{ref:wideresnet_2016}.
We train the networks for 200 epochs using an initial learning rate of 0.1, and decay the learning rate at epochs 75 and 160 by 0.1.
All experiments are trained with stochastic gradient descent (SGD) with a momentum of $0.9$.

\subsection{Long-tailed classification benchmarks}
\label{sub:exp:main_exp}

\subsubsection{CIFAR-100-LT}
\label{subsub:exp:cifar-100}
We conduct experiments on CIFAR-100-LT using different imbalance ratios: 10, 50, 100.
We apply \our to various methods to verify its effectiveness on different algorithms: vanilla cross-entropy loss, class-reweighting loss (LDAM~\cite{ref:cao_ldam_neurips2019}), a two-stage algorithm (DRW~\cite{ref:cao_ldam_neurips2019}), and multi-branch architecture (RIDE~\cite{ref:ride_iclr2021}). 

\noindent\textbf{Comparison with state-of-the-art methods.}
The overall classification accuracies are displayed in Table~\ref{table:cifar100-sota}.
It is surprising that \our with basic cross-entropy (CE) loss shows comparable performance to that of complex long-tail recognition methods.
Moreover, applying \our to the state-of-the-art model (\emph{i.e.}, RIDE) further boosts the performance markedly, especially when the imbalance ratios are high as 50 and 100. 
\begin{table}[]
\caption{
\textbf{State-of-the-art comparison on CIFAR-100-LT dataset.}
Classification accuracy (\%) for ResNet-32 architecture on CIFAR-100-LT with different imbalance ratios.
$\ast$ and $\dagger$ are from the original paper and \cite{ref:lade_cvpr2021}, respectively.
}
\label{table:cifar100-sota}
\centering
\footnotesize
{
\begin{tabular}{lccc}
\toprule
Imbalance ratio         & 100  & 50   & 10   \\ \midrule
Cross Entropy (CE)        & 38.6 & 44.0 & 56.4 \\
CE-DRW        & 41.1 & 45.6 & 57.9  \\
LDAM-DRW~\cite{ref:cao_ldam_neurips2019} & 41.7 & 47.9 & 57.3 \\
BBN~\cite{ref:zhou_chen_bbn_cvpr2020}$^\dagger$                    & 42.6 & 47.1 & 59.2 \\
Causal Norm~\cite{ref:causalnorm_neurips2020}$^\dagger$            & 44.1 & 50.3 & 59.6 \\
IB Loss~\cite{ref:ibloss_iccv2021}$^\ast$      & 45.0 & 48.9 & 58.0 \\
Balanced Softmax (BS)~\cite{ref:ren_bms_neurips2020}$^\dagger$ & 45.1 & 49.9 &61.6 \\
LADE~\cite{ref:lade_cvpr2021}$^\dagger$      & 45.4 & 50.5 & 61.7 \\
Remix~\cite{ref:remix_eccv2020}     & 45.8  & 49.5  & 59.2  \\
RIDE (3 experts)~\cite{ref:ride_iclr2021}        & 48.6 & 51.4 & 59.8 \\ 
MiSLAS ~\cite{ref:mislas_cvpr2021}$^\ast$   & 47.0 & 52.3 & \textbf{63.2} \\ \hline
\rowcolor{lightcolor}
CE + \our                   & 43.9 & 48.3 & 59.5 \\ 
\rowcolor{lightcolor}
CE-DRW + \our                   & 47.0 & 50.9  & 61.7  \\ 
\rowcolor{lightcolor}
LDAM-DRW + \our             & 47.2 & 51.7 & 58.4 \\ 
\rowcolor{lightcolor}
BS + \our                 & 46.6 & 51.4 & 62.3 \\
\rowcolor{lightcolor}
RIDE (3 experts) + \our                 & \textbf{50.0} & \textbf{53.0} & 60.2 \\   \bottomrule
\end{tabular}
\vspace{-2mm}
}
\end{table}

\begin{table}[]
\caption{
\textbf{Comparison against baselines on CIFAR-100-LT (Imbalance ratio = 100).}
Classification accuracy (\%) of ResNet-32.
}
\label{table:cifar100-compare-augs}
\centering
\tabcolsep=0.1cm
\footnotesize
{
\begin{tabular}{lcccc}
\toprule
         & Vanilla                                               & +ROS~\cite{ref:resample_hulse_icml07}                                                    & +Remix~\cite{ref:remix_eccv2020}                                                 & +\our                                                    \\ \midrule
CE       & \begin{tabular}[c]{@{}c@{}}38.6\\ (+0.0)\end{tabular} & \begin{tabular}[c]{@{}c@{}}32.3\\ \textcolor{red}{(-5.3)}\end{tabular}  & \begin{tabular}[c]{@{}c@{}}40.0\\ \textcolor{newgreen}{(+1.4)}\end{tabular} & \begin{tabular}[c]{@{}c@{}}\textbf{43.9}\\ \textbf{\textcolor{newgreen}{(+5.3)}}\end{tabular} \\ \hline
CE-DRW~\cite{ref:cao_ldam_neurips2019}   & \begin{tabular}[c]{@{}c@{}}41.1\\ (+0.0)\end{tabular} & \begin{tabular}[c]{@{}c@{}}35.9\\ \textcolor{red}{(-5.2)}\end{tabular}  & \begin{tabular}[c]{@{}c@{}}45.8\\ \textcolor{newgreen}{(+4.7)}\end{tabular} & \begin{tabular}[c]{@{}c@{}}\textbf{47.0}\\ \textbf{\textcolor{newgreen}{(+5.9)}}\end{tabular} \\\hline
LDAM-DRW~\cite{ref:cao_ldam_neurips2019} & \begin{tabular}[c]{@{}c@{}}41.7\\ (+0.0)\end{tabular} & \begin{tabular}[c]{@{}c@{}}32.6\\ \textcolor{red}{(-9.1)}\end{tabular}  & \begin{tabular}[c]{@{}c@{}}45.3\\ \textcolor{newgreen}{(+3.6)}\end{tabular} & \begin{tabular}[c]{@{}c@{}}\textbf{47.2}\\ \textbf{\textcolor{newgreen}{(+5.5)}}\end{tabular} \\\hline
RIDE~\cite{ref:ride_iclr2021}     & \begin{tabular}[c]{@{}c@{}}48.6\\ (+0.0)\end{tabular} & \begin{tabular}[c]{@{}c@{}}22.6\\ \textcolor{red}{(-26.0)}\end{tabular} & \begin{tabular}[c]{@{}c@{}}44.0\\ \textcolor{red}{(-4.6)}\end{tabular} & \begin{tabular}[c]{@{}c@{}}\textbf{50.0}\\ \textbf{\textcolor{newgreen}{(+1.4)}}\end{tabular} \\ \bottomrule
\end{tabular}
}
\end{table}


\noindent\textbf{Comparison with oversampling methods.}
We further compare the performance improvement of \our with that of other oversampling techniques when combined with long-tailed recognition methods (see Table~\ref{table:cifar100-compare-augs}).
The results reveal that \our consistently improves the performance of all long-tailed recognition methods.
On the other hand, simply balancing the class distribution with ROS~\cite{ref:resample_hulse_icml07} severely degrades performance.
We speculate that this is because the naive balancing of the sampling distribution across classes hinders the model from learning generalized features for major classes and induces the model to memorize the minor class samples.
Remix~\cite{ref:remix_eccv2020} improves the performance of some methods but degrades the performance when combined with RIDE~\cite{ref:ride_iclr2021}.
This indicates that the simple labeling policy of Remix may not be effective when the model complexity becomes large, as in RIDE.

\subsubsection{ImageNet-LT}

\noindent\textbf{Comparison with state-of-the-art methods.}
The results of our method and other long-tailed recognition methods are displayed in Table~\ref{table:imagenet-lt-sota}.
Applying \our to the basic training with CE loss improves the performance by a significant margin, outperforming most of the recent baselines.
The greater performance improvement on ImageNet-LT compared to CIFAR-100 indicates that our method benefits from the richer context information available in the major classes of ImageNet-LT.
In addition, a consistent performance improvement by using \our when combined with DRW or BS bolsters the efficacy of \our, which can be easily integrated into modern state-of-the-art long-tailed recognition methods.
It is noteworthy that as \{CE-DRW + \our\} and \{BS + \our\} especially achieve a much higher few-shot class accuracy than did the other methods, our method is useful for achieving consistent performance across classes.
Lastly, applying CMO to RIDE further boosts performance, outperforming the results of RIDE with four experts.
\\
\begin{table}[]
\caption{
\textbf{State-of-the-art comparison on ImageNet-LT.}
Classification accuracy (\%) of ResNet-50 with state-of-the-art methods trained for 90 or 100 epochs.
``$\ast$" and ``$\dagger$" denote the results are from the original papers, and ~\cite{ref:kang_decoupling_iclr2020}, respectively.
The best results are marked in bold.
}
\label{table:imagenet-lt-sota}
\centering
\tabcolsep=0.2cm
\footnotesize
{
\begin{tabular}{lcccc}
\toprule
                   & All  & Many & Med  & Few  \\ \midrule
Cross Entropy (CE)$^\dagger$ & 41.6 & 64.0 & 33.8 & 5.8  \\
Decouple-cRT~\cite{ref:kang_decoupling_iclr2020}$^\dagger$                & 47.3 & 58.8 & 44.0 & 26.1 \\
Decouple-LWS~\cite{ref:kang_decoupling_iclr2020}$^\dagger$                 & 47.7 & 57.1 & 45.2 & 29.3 \\
Remix~\cite{ref:remix_eccv2020} & 48.6 & 60.4 & 46.9 & 30.7 \\
LDAM-DRW~\cite{ref:cao_ldam_neurips2019} & 49.8 & 60.4 & 46.9 & 30.7 \\
CE-DRW & 50.1 & 61.7 & 47.3 & 28.8 \\
Balanced Softmax (BS)~\cite{ref:ren_bms_neurips2020} & 51.0 & 60.9 & 48.8 & 32.1  \\
Causal Norm~\cite{ref:causalnorm_neurips2020}$^\ast$         & 51.8 & 62.7 & 48.8 & 31.6 \\
RIDE (3 experts)~\cite{ref:ride_iclr2021}$^\ast$            & 54.9 & 66.2 & 51.7 & 34.9 \\ 
RIDE (4 experts)~\cite{ref:ride_iclr2021}$^\ast$            & 55.4 & 66.2 & 52.3 & 36.5 \\ \hline
\rowcolor{lightcolor}
CE + \our            & 49.1 & \textbf{67.0} & 42.3 & 20.5 \\
\rowcolor{lightcolor}
CE-DRW + \our        & 51.4 & 60.8 & 48.6 & 35.5 \\ 
\rowcolor{lightcolor}
\rowcolor{lightcolor}
LDAM-DRW + \our        & 51.1 & 62.0 &	47.4 &	30.8 \\ 
\rowcolor{lightcolor}
BS + \our            & {52.3} & 62.0 & 49.1 & \textbf{36.7} \\
\rowcolor{lightcolor}
RIDE (3 experts) + \our            & \textbf{56.2} & 66.4 & \textbf{53.9} & {35.6} \\ \bottomrule
\end{tabular}
}
\end{table}

\noindent\textbf{Comparison with oversampling methods.}
In Table~\ref{table:imagenet-compare-augs}, we compare performance improvement using other oversampling techniques.
While \our consistently improves performance for all methods, Remix~\cite{ref:remix_eccv2020} fails to improve the performance of the long-tailed recognition methods and barely improves the model trained with cross-entropy loss.
This implies that the labeling strategy of Remix is not sufficient to compensate for the adverse effect of using the same original distribution as the two sampling distributions of the mixup method, especially when the imbalance ratio rises severly to 256, as with ImageNet-LT.
In contrast, \our generates more minority samples by using different distributions when selecting two images and produces much better classification accuracy on all tasks.
\begin{table}[]
\caption{
\textbf{Comparison against baselines on ImageNet-LT.}
Classification accuracy (\%) of ResNet-50.
}
\label{table:imagenet-compare-augs}
\centering
\tabcolsep=0.1cm
\footnotesize
{
\begin{tabular}{lccc}
\toprule
                 & Vanilla                                               & +Remix~\cite{ref:remix_eccv2020}                                                 & +\our                                                  \\ \midrule
CE               & \begin{tabular}[c]{@{}c@{}}41.6\\ (+0.0)\end{tabular} & \begin{tabular}[c]{@{}c@{}}41.7\\ \textcolor{newgreen}{(+0.1)}\end{tabular} & \begin{tabular}[c]{@{}c@{}}\textbf{49.1}\\ \textbf{\textcolor{newgreen}{(+7.5)}}\end{tabular} \\ \hline
CE-DRW~\cite{ref:cao_ldam_neurips2019}           & \begin{tabular}[c]{@{}c@{}}50.1\\ (+0.0)\end{tabular} & \begin{tabular}[c]{@{}c@{}}48.6\\ \textcolor{red}{(-1.5)}\end{tabular} & \begin{tabular}[c]{@{}c@{}}\textbf{51.4}\\ \textbf{\textcolor{newgreen}{(+1.3)}}\end{tabular} \\ \hline
Balanced Softmax~\cite{ref:ren_bms_neurips2020}& \begin{tabular}[c]{@{}c@{}}51.0\\ (+0.0)\end{tabular} & \begin{tabular}[c]{@{}c@{}}49.2\\ \textcolor{red}{(-1.8)}\end{tabular} & \begin{tabular}[c]{@{}c@{}}\textbf{52.3}\\ \textbf{\textcolor{newgreen}{(+1.3)}}\end{tabular} \\ \bottomrule
\end{tabular}
\vspace{-2mm}
}
\end{table}


\noindent\textbf{Results on longer training epochs.}
Recently, PaCo~\cite{ref:paco_iccv2021} performed impressively by using supervised contrastive learning.
Since contrastive learning requires diverse augmentation strategies and longer training times, PaCo trained networks for 400 epochs using RandAugment~\cite{ref:randaug_neurips2020}. 
Since \our should also improve using longer training epochs, we evaluate \our using the same setting from PaCo (\emph{i.e.}, 400 epochs \& RandAug).
Table~\ref{table:imagenet-lt-paco-setting} reveals that \{BS + \our\} achieves a new state-of-the-art performance.
It is noteworthy that applying \our significantly surpasses the two baselines, especially in the few-shot classes.
On top of its simplicity and much lower computational cost, the results demonstrate the effectiveness of the proposed method.
The results on CIFAR-100-LT and iNaturalist 2018 are included in the Supplementary Material.

\begin{table}[]
\caption{
\textbf{Results on longer training epochs with RandAugment~\cite{ref:randaug_neurips2020}.}
Classification accuracy (\%) of ResNet-50 on ImageNet-LT.
``$\ast$" denotes the results from~\cite{ref:paco_iccv2021}.
}
\label{table:imagenet-lt-paco-setting}
\centering
\tabcolsep=0.2cm
\footnotesize
{
\begin{tabular}{lcccc}
\toprule
                   & All  & Many & Med  & Few  \\ \midrule
BS$^\ast$                 & {55.0} & 66.7 & 52.9 & 33.0 \\
PaCo ~\cite{ref:paco_iccv2021}$^\ast$        & 57.0 & 65.0 & \textbf{55.7} & 38.2 \\
\rowcolor{lightcolor}
BS + \our                & \textbf{58.0} & \textbf{67.0} & 55.0 & \textbf{44.2} \\ \bottomrule
\end{tabular}
}
\end{table}

\vspace{-2mm}

\begin{table}[]
\caption{
\textbf{State-of-the-art comparison on iNaturalist2018.}
Classification accuracy (\%) of ResNet-50 on iNaturalist2018.
``$\ast$" and ``$\dagger$" indicate the results from the original paper and \cite{ref:zhou_chen_bbn_cvpr2020}, respectively. RIDE~\cite{ref:ride_iclr2021} was trained for 100 epochs. 
}
\label{table:inat}
\centering
\footnotesize
{
\begin{tabular}{lcccc}
\toprule
               & All  & Many & Med  & Few  \\ \midrule
Cross Entropy (CE) & 61.0 & 73.9 & 63.5 & 55.5 \\
IB Loss~\cite{ref:ibloss_iccv2021}$^\ast$        & 65.4 & -    & -    & -    \\
FSA~\cite{ref:feature_aug_eccv2020}$^\ast$ & 65.9 & - & - & - \\
LDAM-DRW~\cite{ref:cao_ldam_neurips2019}$^\dagger$       & 66.1 & -    & -    & -    \\
Decouple-cRT~\cite{ref:kang_decoupling_iclr2020}$^\ast$            & 68.2 & 73.2 & 68.8 & 66.1 \\
Decouple-LWS~\cite{ref:kang_decoupling_iclr2020}$^\ast$            & 69.5 & 71.0 & 69.8 & 68.8 \\
BBN~\cite{ref:zhou_chen_bbn_cvpr2020}$^\ast$            & 69.6 & -    & -    & -    \\
Balanced Softmax~\cite{ref:ren_bms_neurips2020}         & 70.0 & 70.0    & 70.2    & 69.9    \\
LADE~\cite{ref:lade_cvpr2021}$^\ast$           & 70.0 & -    & -    & -    \\
Remix~\cite{ref:remix_eccv2020}$^\ast$ & 70.5 & -    & -    & -    \\ 
MiSLAS~\cite{ref:mislas_cvpr2021}$^\ast$ & 71.6 & 73.2 & 72.4 & 70.4 \\ 
RIDE (3 experts)~\cite{ref:ride_iclr2021}$^\ast$ & 72.2 & 70.2 & 72.2 & 72.7 \\ 
RIDE (4 experts)~\cite{ref:ride_iclr2021}$^\ast$ & 72.6 & 70.9 & 72.4 & \textbf{73.1} \\ \hline
\rowcolor{lightcolor}
CE + \our  & 68.9 & \textbf{76.9} & 69.3 & 66.6 \\
\rowcolor{lightcolor}
CE-DRW + \our  & 70.9 & 68.2 & 70.2 & 72.2 \\
\rowcolor{lightcolor}
LDAM-DRW + \our  & 69.1 & 75.3 & 69.5 & 67.3 \\
\rowcolor{lightcolor}
BS + \our  & 70.9 & 68.8 & 70.0 & {72.3} \\
\rowcolor{lightcolor}
CE-DRW + \our + LAS~\cite{ref:mislas_cvpr2021} & {71.8}  & 69.6 & {72.1}  &71.9 \\ 
\rowcolor{lightcolor}
RIDE (3 experts) + \our  & \textbf{72.8} & 68.7	& \textbf{72.6} & \textbf{73.1} \\ \bottomrule
\end{tabular}
}
\end{table}


\begin{figure*}[t]
\centering{%
\includegraphics[width=1\linewidth]{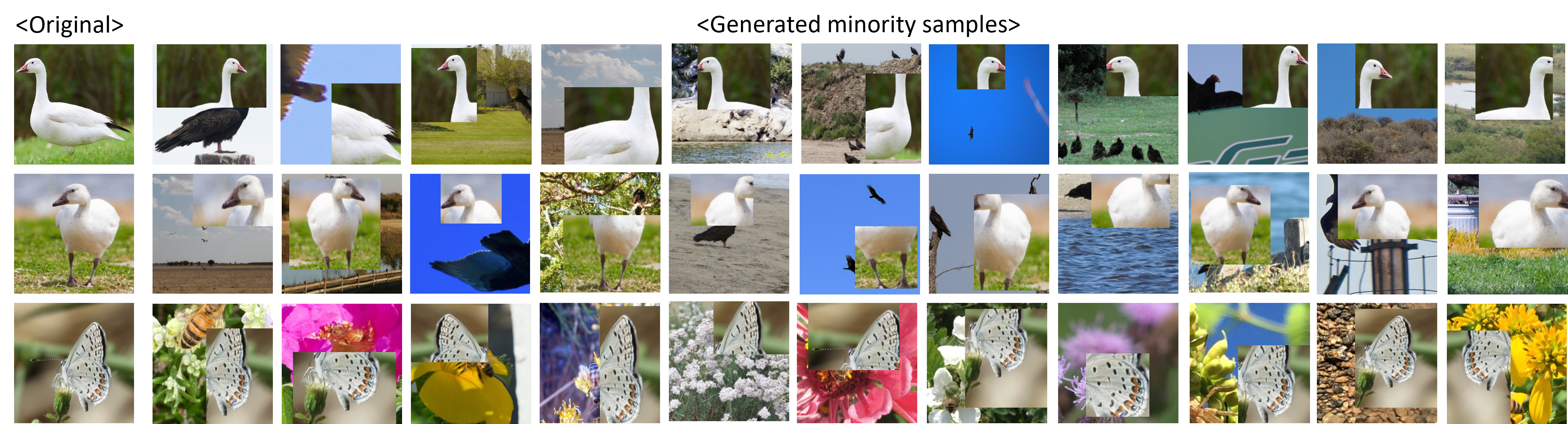}}
\caption{\textbf{A display of the minority images generated by \our}
(minority classes: the snow goose and the Acmon blue (butterfly)). We randomly choose generated images for each original image. Our method is able to generate context-rich minority samples that have diverse contexts. 
For example, while the original `snow goose' class contains only images of a `snow goose' on grass, the generated images have various contexts such as the sky, the sea, the sand, and a flock of crows.
These generated images enable the model to learn a robust representation of minority classes.}
\label{fig:generated_images}
\vspace{-7mm}
\end{figure*}

\subsubsection{iNaturalist 2018}
\noindent\textbf{Comparison with state-of-the-art methods.}
Table~\ref{table:inat} presents the classification results.
On the naturally-skewed dataset, applying \our to the simple training scheme of CE-DRW surpasses most of the state-of-the-arts.
On iNaturalist 2018, as in ImageNet-LT, \our dramatically improves the performance of the cross-entropy loss (CE) by \textbf{7.9\%p} ($61.0\%$ increased to $68.9\%$).
This is because the sample generation by \our fully utilizes the abundant context of training data.
Again, it achieves a remarkable performance improvement in the few-shot classes.
It is moreover noteworthy that when we apply the same stage-2 strategy, LAS, from ~\cite{ref:mislas_cvpr2021}, it further boosts performance.
Lastly, applying \our to RIDE achieves a new state-of-the-art performance.\\

\noindent\textbf{Results on large models.}
We investigate the performance of \our and other oversampling methods using the large deep networks of Wide ResNet-50~\cite{ref:wideresnet_2016}, ResNet-101, and ResNet-152~\cite{ref:resnet_cvpr2016}.
We compare \our with the feature space augmentation method (FSA)~\cite{ref:feature_aug_eccv2020}.
While both methods improve the results from vanilla training with cross-entropy loss, our method provides superior performance to that of FSA.
This indicates that using the context-rich information from majority classes in the input space is simple but effective in improving the overall performance.

\begin{table}[h]
\caption{
\textbf{Results on large architectures.}
Classification accuracy (\%) of large backbone networks on iNaturalist 2018.
The results are copied from~\cite{ref:feature_aug_eccv2020}.
}
\label{table:inat-diff-archs}
\centering
\tabcolsep=0.09cm
\footnotesize
{
\begin{tabular}{lcccc}
\toprule
Method & ResNet-50 & Wide ResNet-50 & ResNet-101 & ResNet-152 \\ \midrule
CE & 61.0 & - & 65.2 & 66.2 \\
FSA~\cite{ref:feature_aug_eccv2020}   & 65.9      & -              & 68.4       & 69.1       \\
\rowcolor{lightcolor}
\our  & \textbf{70.9}      & \textbf{71.9}           & \textbf{72.4}       & \textbf{72.6}       \\ \bottomrule
\end{tabular}
\vspace{-2mm}
}
\end{table}

\noindent\textbf{Display of the generated images.}
We visualize the generated images for the minority classes in Figure~\ref{fig:generated_images}.
From the rarest minority classes, we randomly choose generated images for each original image.
\our produces diverse minority samples that have various contexts.
For example, while the `snow goose' class contains only images of geese on grass, the generated images have various contexts, such as the sky or sea.
Likewise, the butterflies in the third row are newly created as diverse images that have various contexts, containing bees and flowers of various colors and shapes.
We argue that various combinations of context and minority samples encourage the model to learn a robust representation of the minority classes.

\subsection{Analysis} 
\label{sub:exp_analysis}

\noindent\textbf{Is the distribution for augmenting images important? }
To justify the need for different distributions of background and foreground images, we compare CutMix and \our.
As can be seen from Table~\ref{table:compare-cutmix-our}, \our outperforms CutMix on long-tailed classification by a large margin.
In particular, there is a remarkable performance improvement in the medium and few-shot classes.
The performance gap is due to the absence of a minor-class-weighted distribution in CutMix augmentation. 
Although CutMix can generate informative mixed samples, its effect is limited when used with long-tailed distributions.
Thus, we claim that the use of a  minor-class-weighted distribution is a key-point in data augmentation in the long-tailed settings; this highlights the contribution and originality of \our. \\
\vspace{-2mm}
\begin{table}[h]
\caption{
\textbf{Comparison with CutMix}
using cross-entropy loss.
}
\label{table:compare-cutmix-our}
\centering
\footnotesize
{
\begin{tabular}{lcccc}
\toprule
                    & All & Many & Med & Few \\ \midrule
\textit{\textbf{CIFAR-100-LT}}           &     &      &     &     \\
CutMix              & 35.6   & 71.0    & 37.9   & 4.9   \\
\rowcolor{lightcolor}
\our                & 43.9   & 70.4    & 42.5   & 14.4   \\ \midrule
\textit{\textbf{ImageNet-LT}}       &     &      &     &     \\
CutMix              & 45.5   & 68.6    & 38.1   & 8.1   \\
\rowcolor{lightcolor}
\our                & 49.1   & 67.0    & 42.3   & 20.5   \\
\bottomrule
\end{tabular}
\vspace{-1mm}
}
\end{table}

\noindent\textbf{How to choose the appropriate probability distribution $Q$.}
We evaluate different sampling strategies in Section~\ref{sub:method_q} on CIFAR-100 with the imbalance ratio 100,
The results are reported in Table~\ref{table:res-dist-cifar100}.
$q(1, k)$ displays the most balanced performance.
This result is consistent with the common practice of balancing the dataset by assigning weights inversely proportional to the class frequency.
While $q(2, k)$, which imposes a higher probability on the minority class than does $q(1, k)$, performs acceptably in the few-shot classes, the overall performance slightly deteriorates.
We assume this is because we cannot sample more diverse images when imposing too high probabilities on the few-shot classes.
Based on this result, we set $Q$ as $q(1, k)$ in our all experiments.
\begin{table}[h]
\caption{
\textbf{Impact of different $Q$ sampling distributions.}
Results on CIFAR-100-LT (imbalance ratio=100) according to different $Q$ sampling probabilities.
}
\label{table:res-dist-cifar100}
\centering
\footnotesize
{
\begin{tabular}{lcccc}
\toprule
         & All & Many & Med & Few \\ \midrule
$q(1/2,k)$ & 42.6   & \textbf{71.6}    & 42.1   & 9.5   \\
\rowcolor{lightcolor}
$q(1,k)$   & \textbf{43.9}   & 70.4    & \textbf{42.5}   & \textbf{14.4}   \\
$q(2,k)$   & 40.1   & 67.2    & 36.7   & 12.3   \\
$E(k)$~\cite{ref:cui_belongie_cvpr19}     & 39.5   & 70.4    & 38.0   & 4.7   \\ \bottomrule
\end{tabular}
}
\end{table}

\vspace{2mm}
\noindent\textbf{Why should we oversample only for the foreground samples?}
One may wonder why oversampling only for the foreground samples is better than oversampling both patches and background samples or oversampling only the backgrounds.
To verify our design choice, we evaluate two variants of \our.
The first variant, \our$_{back}$, samples background images from a minor-class-weighted distribution and patches from the original distribution, which is exactly the opposite design of \our, \emph{\emph{i.e.}}, $(x^b, y^b) \sim Q, (x^f, y^f) \sim P$.
The second variant, \our$_{minor}$, samples both the background and the patches from a  minor-class-weighted distribution, \emph{i.e.}, $(x^b, y^b), (x^f, y^f) \sim Q$.
We report the results of applying these variants of the \our method to the model trained with CE loss and LDAM loss~\cite{ref:cao_ldam_neurips2019} in Table~\ref{table:compare-variants}.
\begin{table}[h]
\caption{
\textbf{Ablation study.}
Results from variants of \our with ResNet-32 on imbalanced CIFAR-100; imbalance ratio of 100.
}
\label{table:compare-variants}
\centering
\footnotesize
{
\begin{tabular}{lcccc}
\toprule
                   & All & Many & Med & Few \\ \midrule
Cross Entropy (CE)                  & 38.6   & 65.3    & 37.6   & 8.7   \\
CE + \our$_{minor}$                 & 37.9   & 58.3    & 40.4   & 11.2   \\
CE + \our$_{back}$                  & 40.1   & 64.7    & 40.2   & 11.3   \\
\rowcolor{lightcolor}
CE + \our                           & \textbf{43.9}   & \textbf{70.4}    & \textbf{42.5}   & \textbf{14.4}   \\ \midrule
LDAM~\cite{ref:cao_ldam_neurips2019}& 41.7   & 61.4    & 42.2   & 18.0   \\
LDAM + \our$_{minor}$               & 31.7   & 50.2    & 33.2    & 8.4    \\
LDAM + \our$_{back}$                & 44.2   & 59.2    & 46.6   & 24.0   \\
\rowcolor{lightcolor}
LDAM + \our                         & \textbf{47.2}   & \textbf{61.5}    & \textbf{48.6}   & \textbf{28.8}   \\ \bottomrule
\end{tabular}
}
\end{table}

\our$_{minor}$ yields severe performance degradation using both methods.
We suspect that this is because the rich context of the majority samples cannot be utilized.
In contrast, \our$_{back}$ produces acceptable performance improvements, but far less than did the original \our.
This is because, using the CutMix method, there is a high probability that the object in the foreground image overlaps the background image.
Therefore, we can expect a loss of information about minority classes in the background image, resulting in a limited performance boost. \\


\noindent\textbf{Comparison with other minority augmentations.}
To further verify our design choice, we analyze the effectiveness of using different augmentation methods, including CutMix~\cite{ref:cutmix_iccv2019}, Mixup~\cite{ref:mixup_iclr2018}, color jitter, and Gaussian blur.
For Mixup, we use the same sampling strategy as for \our.
For color jitter and Gaussian blur, which do not interpolate two images, we apply augmentation only to the minority classes and oversample those classes. 
As evidenced in Table~\ref{table:compare-augs}, other augmentation methods provide little performance gain compared to the gains using CutMix.
We suspect that this is because the pixel-level transformations (\emph{i.e.}, Gaussian blur and color jitter) are not effective in producing minority samples that have a rich context.
Gaussian blur and color jitter do not combine two images; thus, it is hard to add a new context to minority samples.
While Mixup combines two images, it does not distinguish the roles of the two samples, limiting the control of the source of the context and of the patch information.
In contrast, CutMix can create diverse images with larger changes at pixel-level than can other methods.
\begin{table}[h]
\caption{
\textbf{Data augmentation methods.}
Comparisons between augmentation methods for generating new minority samples on CIFAR-100-LT with an imbalance ratio of 100.
}
\label{table:compare-augs}
\centering
\footnotesize
{
\begin{tabular}{lcccc}
\toprule
                                                   & All & Many & Med & Few \\ \midrule
\our w/ Gaussian Blur                                    & 31.1   & 54.7    & 28.8   & 6.2   \\
\our w/ Color Jitter                                  & 34.7   & 58.9    & 34.4   & 6.8   \\
\our w/ Mixup             & 38.0   & 54.8    & 40.2   & \textbf{15.9}   \\
\rowcolor{lightcolor}
\our w/ CutMix                                              & \textbf{43.9}   & \textbf{70.4}    & \textbf{42.5}   & 14.4   \\ \bottomrule
\end{tabular}
}
\end{table}







\section{Conclusion}
\label{sec:con}
We have proposed a novel {context-rich} oversampling method, \our, to solve the data imbalance problem.
We tackle the fundamental problem of previous oversampling methods that generate {context-limited} minority samples, which intensifies the overfitting problem. 
Our key idea is to transfer the rich contexts of majority samples to minority samples to augment minority samples. 
The implementation of \our is simple and intuitive.
Extensive experiments on various benchmark datasets demonstrate not only that our \our significantly improves performance, but also that adding our oversampling method to the basic losses advances the state-of-the-art.


{
\noindent\textbf{Limitations.}
In some cases, the performance improvement for the minority classes occurs with the degraded performance of the majority classes.
Future work should be designed to improve the performance of all classes without sacrificing the performance of the many-shot classes. 

\noindent\textbf{Potential negative societal impact.}
Since our method creates new samples, it benefits more from longer training and deeper architectures.
Thus, it may lead to more computations, which has a risk that the use of GPUs for machine learning could accelerate environmental degradation~\cite{ref:discussion_earth}.

{
\noindent\textbf{Acknowledgement}
This work was partially supported by IITP grant from Korea government (MSIT) [No.B0101-15-0266, 
High Performance Visual BigData Discovery Platform;
NO.2021-0-01343, AI Graduate School Program (SNU)]}

}
\clearpage
{\small
\bibliographystyle{ieee_fullname}
\bibliography{egbib}
}
\newpage
\appendix
\onecolumn
\renewcommand*{\thesection}{\Alph{section}}
\renewcommand{\theequation}{\thesection.\arabic{equation}}

\begin{center}
\bf {\LARGE Supplementary Material}
\end{center}

\section{Implementation details}\label{sup:implementation}

\noindent In this section, we provide implementation details that are not included in Section 4.1. \\

\noindent\textbf{CIFAR-100-LT.}
To set up a fair comparison, we used the same random seed to make CIFAR-100-LT, and followed the implementation of ~\cite{ref:cao_ldam_neurips2019}.
We trained ResNet-32~\cite{ref:resnet_cvpr2016} by SGD optimizer with a momentum of 0.9, and a weight decay of $2 \times 10^{-4}$.
As in ~\cite{ref:cao_ldam_neurips2019}, we used simple data augmentation~\cite{ref:resnet_cvpr2016} by padding 4 pixels on each side and applying horizontal flipping or random cropping to $32\times32$ size.
We trained for 200 epochs and used a linear warm-up of the learning rate~\cite{ref:goyal_he_corr2017} in the first five epochs. 
The learning rate was initialized as 0.1, and it was decayed at the 160th and 180th epoch by 0.01.
The model was trained with a batch size of 128 on a single GTX 1080Ti.
We turned off \our for the last three epochs in order to finetune the model in the original input space.

\noindent For experiments in Table 11, we use the same strategy as for \{\our w/ Mixup\}.
For \{\our w/ Gaussian Blur\} and \{\our w/ Color Jitter\}, which do not mix two images, we divided classes into two groups: the majority and the minority.
Then, for the minority group, we augmented the data with color jitter and gaussian blur, respectively.
We set brightness to 0.5 and hue to 0.3 for color jitter, and set kernel size as $(5, 7)$ and sigma as $(0.1, 5)$ for Gaussian blur using the PyTorch~\cite{ref:Pytorch} implemented functions. \\

\noindent\textbf{ImageNet-LT.}
For ImageNet-LT, we followed most of the details from ~\cite{ref:ride_iclr2021}.
As in ~\cite{ref:ride_iclr2021}, we performed simple horizontal flips, color jittering, and took random crops $224 \times 224$ in size.
We used ResNet-50 as a backbone network.
The networks were trained with a batch size of 256 on 4 GTX 1080Ti GPUs for 100 epochs using SGD and an initial learning rate of 0.1; this rate decayed by 0.1 at 60 epochs and 80 epochs.\\

\noindent\textbf{iNaturalist 2018.}
For iNaturalist 2018, we used the same data augmentation method as for ImageNet-LT.
Multiple backbone networks were experimented on iNaturalist 2018, including ResNet-50, ResNet-101, ResNet-152~\cite{ref:resnet_cvpr2016}, and Wide ResNet-50~\cite{ref:wideresnet_2016}. 
All backbone networks were trained with a batch size of 512 on 8 Tesla V100 GPUs for 200 epochs using SGD at an initial learning rate of 0.1; this rate decayed by 0.1 at 75 epochs and 160 epochs.
Experiments were implemented and evaluated on the NAVER Smart Machine Learning (NSML)~\cite{ref:nsml} platform.

\section{Ablation studies}
\subsection{Comparison with oversampling methods}
\noindent We compare \our with other oversampling methods for performance improvement on CIFAR-100 with imbalance ratio 50 and 10 in Table~\ref{sup:table:cifar100-augs}.
As in the imbalance ratio of 100, our method consistently improves performance in all long-tailed recognition methods.

\begin{table}[h]
\caption{
\textbf{Comparison against baselines on CIFAR-100-LT}
Results with classification accuracy (\%) of ResNet-32. The best results are marked in bold.
}
\label{sup:table:cifar100-augs}
\centering
\small
{
\begin{tabular}{l|cccc|cccc}
\toprule
Imbalance ratio & \multicolumn{4}{c|}{50}                                                                                                                                                                                                        & \multicolumn{4}{c}{10}                                                                                                                                                                                                         \\ \hline
Method          & Vanilla                                               & +ROS~\cite{ref:resample_hulse_icml07}                                                   & +Remix~\cite{ref:remix_eccv2020}                                                & +CMO                                                  & Vanilla                                               & +ROS~\cite{ref:resample_hulse_icml07}                                                   & +Remix~\cite{ref:remix_eccv2020}                                                & +CMO                                                  \\ \midrule
CE              & \begin{tabular}[c]{@{}c@{}}44.0\\ (+0.0)\end{tabular} & \begin{tabular}[c]{@{}c@{}}39.7\\ \textcolor{red}{(-4.3)}\end{tabular}  & \begin{tabular}[c]{@{}c@{}}45.0\\ \textcolor{newgreen}{(+1.0)}\end{tabular} & \begin{tabular}[c]{@{}c@{}}\textbf{48.3}\\ \textbf{\textcolor{newgreen}{(+4.3)}}\end{tabular} & \begin{tabular}[c]{@{}c@{}}56.4\\ (+0.0)\end{tabular} & \begin{tabular}[c]{@{}c@{}}55.6\\ \textcolor{red}{(-0.8)}\end{tabular}  & \begin{tabular}[c]{@{}c@{}}58.7\\ \textcolor{newgreen}{(+2.3)}\end{tabular} & \begin{tabular}[c]{@{}c@{}}\textbf{59.5}\\ \textbf{\textcolor{newgreen}{(+3.1)}}\end{tabular} \\ \hline
CE-DRW~\cite{ref:cao_ldam_neurips2019}          & \begin{tabular}[c]{@{}c@{}}45.6\\ (+0.0)\end{tabular} & \begin{tabular}[c]{@{}c@{}}41.3\\ \textcolor{red}{(-4.3)}\end{tabular}  & \begin{tabular}[c]{@{}c@{}}49.5\\ \textcolor{newgreen}{(+3.9)}\end{tabular} & \begin{tabular}[c]{@{}c@{}}\textbf{50.9}\\ \textbf{\textcolor{newgreen}{(+5.3)}}\end{tabular} & \begin{tabular}[c]{@{}c@{}}57.9\\ (+0.0)\end{tabular} & \begin{tabular}[c]{@{}c@{}}56.4\\ \textcolor{red}{(-1.5)}\end{tabular}  & \begin{tabular}[c]{@{}c@{}}59.2\\ \textcolor{newgreen}{(+1.3)}\end{tabular} & \begin{tabular}[c]{@{}c@{}}\textbf{61.7}\\ \textbf{\textcolor{newgreen}{(+3.8)}}\end{tabular} \\ \hline
LDAM-DRW~\cite{ref:cao_ldam_neurips2019}        & \begin{tabular}[c]{@{}c@{}}47.9\\ (+0.0)\end{tabular} & \begin{tabular}[c]{@{}c@{}}38.3\\ \textcolor{red}{(-9.6)}\end{tabular}  & \begin{tabular}[c]{@{}c@{}}48.8\\ \textcolor{newgreen}{(+0.9)}\end{tabular} & \begin{tabular}[c]{@{}c@{}}\textbf{51.7}\\ \textbf{\textcolor{newgreen}{(+3.8)}}\end{tabular} & \begin{tabular}[c]{@{}c@{}}57.3\\ (+0.0)\end{tabular} & \begin{tabular}[c]{@{}c@{}}53.9\\ \textcolor{red}{(-3.4)}\end{tabular}  & \begin{tabular}[c]{@{}c@{}}55.9\\ \textcolor{red}{(-1.4)}\end{tabular} & \begin{tabular}[c]{@{}c@{}}\textbf{58.4}\\ \textbf{\textcolor{newgreen}{(+1.1)}}\end{tabular} \\ \hline
RIDE~\cite{ref:ride_iclr2021}            & \begin{tabular}[c]{@{}c@{}}51.4\\ (+0.0)\end{tabular} & \begin{tabular}[c]{@{}c@{}}31.3\\ \textcolor{red}{(-20.1)}\end{tabular} & \begin{tabular}[c]{@{}c@{}}47.9\\ \textcolor{red}{(-3.5)}\end{tabular} & \begin{tabular}[c]{@{}c@{}}\textbf{53.0}\\ \textbf{\textcolor{newgreen}{(+1.6)}}\end{tabular} & \begin{tabular}[c]{@{}c@{}}59.8\\ (+0.0)\end{tabular} & \begin{tabular}[c]{@{}c@{}}49.4\\ \textcolor{red}{(-10.4)}\end{tabular} & \begin{tabular}[c]{@{}c@{}}59.5\\ \textcolor{red}{(-0.3)}\end{tabular} & \begin{tabular}[c]{@{}c@{}}\textbf{60.2}\\ \textbf{\textcolor{newgreen}{(+0.4)}}\end{tabular} \\ \bottomrule
\end{tabular}
}
\end{table}

\subsection{Results on longer training epochs}
\noindent We evaluate \our using the same setting from PaCo~\cite{ref:paco_iccv2021}.
That is, we train the network for 400 epochs and use AutoAugment~\cite{ref:autoaug_cvpr2019} on CIFAR-100-LT. 
For iNaturalist2018, RandAugment~\cite{ref:randaug_neurips2020} is applied.
Table~\ref{tab:cifar-paco}, \ref{tab:inat-paco} reveals that \{BS + CMO\} surpasses PaCo in most cases, and achieves a new state-of-the-art performance.
These results demonstrate the effectiveness of \our, despite its simplicity.

\vspace{5mm}
\begin{minipage}[c]{0.45\textwidth}
\centering
\small
{
\begin{tabular}{lccc}
\toprule
Imbalance ratio         & 100  & 50   & 10   \\ \midrule
BS$^\ast$                 & {50.8} & 54.2 & 63.0 \\
PaCo ~\cite{ref:paco_iccv2021}$^\ast$        & \textbf{52.0} & {56.0} & {64.2} \\
\rowcolor{lightcolor}
BS + \our                & {51.7} & \textbf{56.7} & \textbf{65.3} \\ \bottomrule
\end{tabular}
}
\captionof{table}{\textbf{Classification Accuracy on CIFAR-100-LT with different imbalance ratios.}
We train ResNet-32 with AutoAugment~\cite{ref:autoaug_cvpr2019} in 400 epochs.
$\ast$ is from ~\cite{ref:paco_iccv2021}
The best results are marked in bold.}
\label{tab:cifar-paco}
\end{minipage} \hspace{6mm}
\begin{minipage}[c]{0.45\textwidth}
\centering
\small
{
\begin{tabular}{lcccc}
\toprule
               & All  & Many & Med  & Few  \\ \midrule
BS$^\ast$                 & 71.8 & \textbf{72.3} & 72.6 & 71.7 \\
PaCo ~\cite{ref:paco_iccv2021}$^\ast$        & 73.2 & 70.3 & 73.2 & 73.6 \\
\rowcolor{lightcolor}
BS + \our                & \textbf{74.0} & 71.9 & \textbf{74.2} & \textbf{74.2} \\ \bottomrule
\end{tabular}
}
\captionof{table}{\textbf{Classification Accuracy on iNaturalist2018.}
We train ResNet-50 for 400 epochs with RandAugment~\cite{ref:randaug_neurips2020}.
``$\ast$" indicates the results are from \cite{ref:paco_iccv2021}.
The best results are marked in bold.}
\label{tab:inat-paco}
\end{minipage}

\subsection{Impact of $\alpha$}
\noindent We evaluate the impact of the hyperparameter $alpha$ in Figure~\ref{fig:alpha_impact}.
The classification accuracy according to different $\alpha \in \{0.1, 0.25, 0.5, 1.0, 2.0, 4.0 \}$ is plotted.
\our improves the baseline accuracy (38.6\%) in all cases.
The best performance is achieved when $\alpha = 1.0$.


\subsection{Computational cost}
\noindent One of the biggest advantages of our method is its low computational cost.
\our only requires to load an additional batch of data from the minor-class-weighted loader.
We measure the training time per batch on ImageNet-LT (see Table \ref{tab:trainingcost}). 
While CE takes \textbf{0.355s}, CE+CMO takes \textbf{0.369s}, which is only an increase of 3.94\%. 

\vspace{5mm}
\begin{minipage}[c]{0.45\textwidth}
\centering
{
\includegraphics[width=0.8\columnwidth]{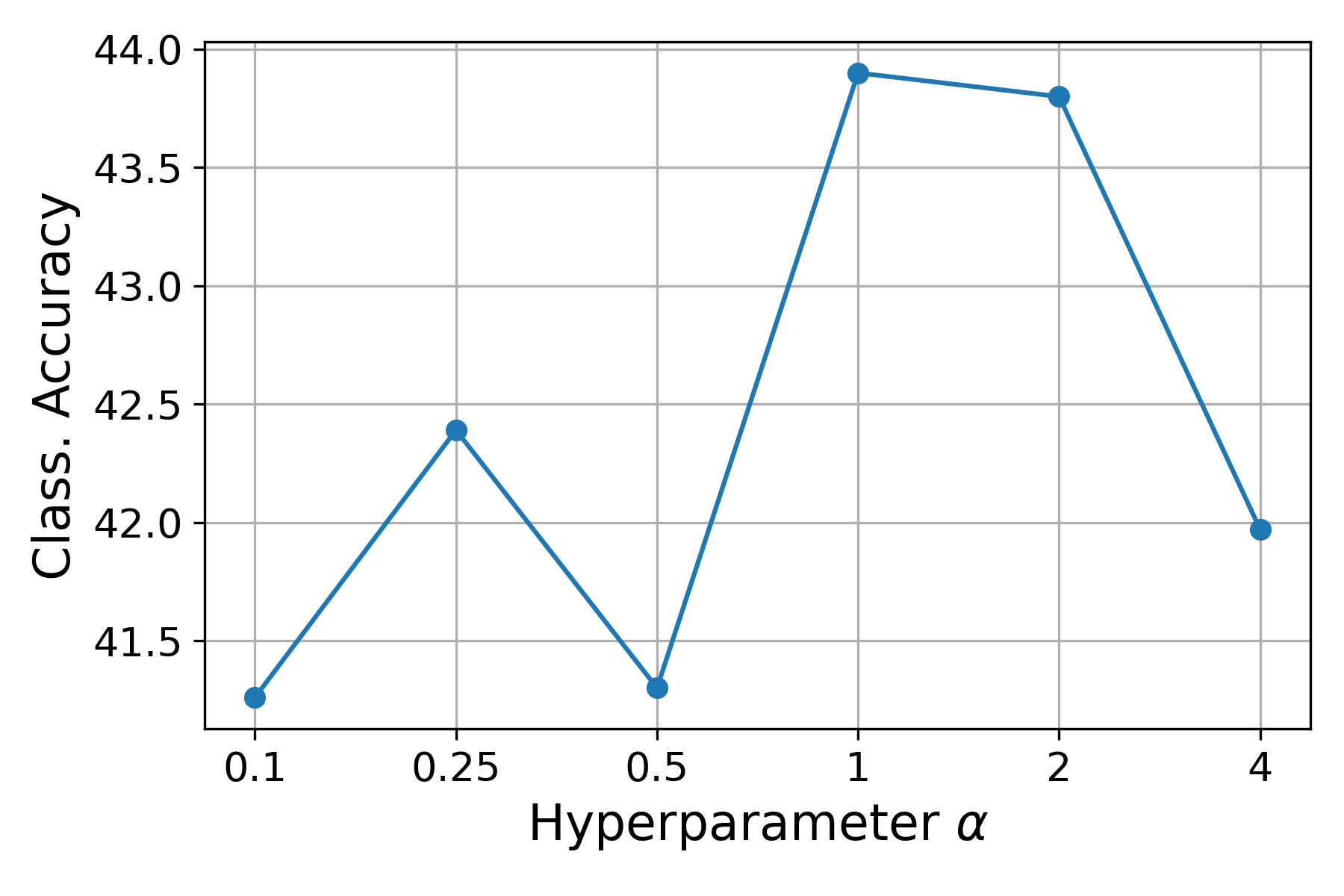}
}
\captionof{figure}{\textbf{Impact of $\alpha$} on CIFAR-100-LT with an imbalance ratio of 100.}
\label{fig:alpha_impact}
\end{minipage} \hspace{6mm}
\begin{minipage}[c]{0.45\textwidth}
\centering
\small
{
\begin{tabular}{lcc}
\toprule
                  & CE    & CE + CMO \\ \midrule
Training Time (s) & 0.355 & 0.369    \\ \bottomrule
\end{tabular}
}
\captionof{table}{\textbf{Training time on ImageNet-LT.}}
\label{tab:trainingcost}
\end{minipage}

\newpage
\section{Pseudo-code of \ourfull}
\noindent We present the PyTorch-syle pseudo-code of \our algorithm in Algorithm~\ref{sup:pytorch-code}.
Note that \our is easy to implement with just a few lines that are easily applicable to any loss, networks, or algorithms. 
Thus, \our can be a very practical and effective solution for handling imbalanced dataset. 

\begin{algorithm}[h]
\caption{PyTorch-style pseudo-code for \our}\label{sup:pytorch-code}
\begin{algorithmic}
\State \PyComment{original\_loader: data loader from original data distribution}
\State \PyComment{weighted\_loader: data loader from minor-class-weighted distribution}
\State \PyComment{model: any backbone network such as ResNet or multi-branch networks (RIDE)}
\State \PyComment{loss: any loss such as CE, LDAM, balanced softmax, RIDE loss} \\
\State \PyCode{for epoch in Epochs:}
\State \hskip1.8em \PyComment{load a batch for background images from original data dist.}
\State \hskip1.8em \PyCode{for x\_b, y\_b in original\_loader:}
\State \hskip4.2em \PyComment{load a batch for foreground from minor-class-weighted dist.}
\State \hskip4.2em \PyCode{x\_f, y\_f = next(weighted\_loader)}\\
\State \hskip4.2em \PyComment{get coordinate for random binary mask}
\State \hskip4.2em \PyCode{lambda = np.random.uniform(0,1)}
\State \hskip4.2em \PyCode{cx = np.random.randint(W)} \PyComment{W: width of images}
\State \hskip4.2em \PyCode{cy = np.random.randint(H)} \PyComment{H: height of images}
\State \hskip4.2em \PyCode{bbx1 = np.clip(cx - int(W * np.sqrt(1. - lambda))//2,0,W)} 
\State \hskip4.2em \PyCode{bbx2 = np.clip(cx + int(W * np.sqrt(1. - lambda))//2,0,W)} 
\State \hskip4.2em \PyCode{bby1 = np.clip(cy - int(H * np.sqrt(1. - lambda))//2,0,H)} 
\State \hskip4.2em \PyCode{bby2 = np.clip(cy + int(H * np.sqrt(1. - lambda))//2,0,H)} \\
\State \hskip4.2em \PyComment{get minor-oversampled images}
\State \hskip4.2em \PyCode{x\_b[:, :, bbx1:bbx2, bby1:bby2] = x\_f[:, :, bbx1:bbx2, bby1:bby2]} 
\State \hskip4.2em \PyCode{lambda = 1 - ((bbx2 - bbx1) * (bby2 - bby1) / (W * H))}\PyComment{adjust lambda} \\
\State \hskip4.2em \PyComment{output (x\_f is attached to x\_b)}
\State \hskip4.2em \PyCode{output = model(x\_b)} \\
\State \hskip4.2em \PyComment{loss}
\State \hskip4.2em \PyCode{losses = loss(output,  y\_b) * lambda + loss(output,  y\_f) * (1. - lambda)} \\
\State \hskip4.2em \PyComment{optimization step}
\State \hskip4.2em \PyCode{losses.backward()}
\State \hskip4.2em \PyCode{optimizer.step()}

\end{algorithmic}
\end{algorithm}



\end{document}